\documentclass[runningheads]{llncs}
\usepackage{amssymb}
\usepackage{booktabs}
\usepackage{multirow}
\usepackage{graphicx}
\usepackage{pdflscape}
\usepackage{amsmath}
\usepackage{amsfonts}
\usepackage{xcolor}
\usepackage{url}
\usepackage{hyperref}
\usepackage{rotating}
\usepackage{float}

\begin{document}
\setlength{\abovedisplayskip}{10pt}
\setlength{\belowdisplayskip}{10pt}
\title{Explainable bank failure prediction models: Counterfactual explanations to reduce the failure risk}
\titlerunning{Explainable bank failure prediction models}
%
\author{Seyma Gunonu \and
Gizem Altun \and
\href{https://orcid.org/0000-0001-8423-1823}{\includegraphics[scale=0.06]{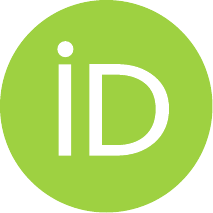}\hspace{1mm}Mustafa Cavus}\inst{*}}
%
\authorrunning{Gunonu et al.}
\institute{Eskisehir Technical University, Department of Statistics, Turkiye \\
\email{*mustafacavus@eskisehir.edu.tr}}
%
\maketitle              

\begin{abstract}
The accuracy and understandability of bank failure prediction models are crucial. While interpretable models like logistic regression are favored for their explainability, complex models such as random forest, support vector machines, and deep learning offer higher predictive performance but lower explainability. These models, known as black boxes, make it difficult to derive actionable insights. To address this challenge, using counterfactual explanations is suggested. These explanations demonstrate how changes in input variables can alter the model's output and suggest ways to mitigate bank failure risk. The key challenge lies in selecting the most effective method for generating useful counterfactuals, which should demonstrate validity, proximity, sparsity, and plausibility. The paper evaluates several counterfactual generation methods: What-If, Multi-Objective, and Nearest Instance Counterfactual Explanation, and also explores resampling methods like undersampling, oversampling, SMOTE, and the cost-sensitive approach to address data imbalance in bank failure prediction in the US. The results indicate that the Nearest Instance Counterfactual Explanation method yields higher-quality counterfactual explanations, mainly using the cost-sensitive approach. Overall, the Multi-Objective Counterfactual and Nearest Instance Counterfactual Explanation methods outperform others regarding validity, proximity, and sparsity metrics, with the cost-sensitive approach providing the most desirable counterfactual explanations. These findings highlight the variability in the performance of counterfactual generation methods across different balancing strategies and machine learning models, offering valuable strategies to enhance the utility of black-box bank failure prediction models.

\end{abstract}

\newpage
\section{Introduction}

Banking plays a significant role in the economic system, and any issues within this sector can have a more substantial adverse impact on the economic system than expected \cite{kristof_and_virag_2022}. Even the failure of a single bank can trigger a chain reaction, potentially spreading swiftly and adversely affecting other banks \cite{le_and_viviani_2018}. Therefore, predicting problems within the banking system in advance has become an essential concern for the scientific community. One such effort is bank failure prediction models employed to forecast the likelihood of a bank's failure using its financial indicators. Like in many other fields, the logistic regression model has been the baseline in bank failure prediction models for many years. This is due to its ease of creation and inherently interpretable structure, making it easily understandable by stakeholders. However, the non-linear complex relationships among variables in the models led to the adoption of more complicated models as alternatives to logistic regression.
Consequently, decision trees, support vector machines, and deep learning models started to be utilized in bank failure prediction models \cite{kristof_and_virag_2022,gogas_et_al_2018,carmona_et_al_2019,petropoulos_et_al_2020,manthoulis_et_al_2021}. Despite exhibiting high prediction performance, these models were overshadowed by their lack of inherently interpretable structures. This is because laws such as GDPR\footnote{Regulation (EU) 2016/679 of the European Parliament and of the Council of 27 April 2016 on the protection of natural persons concerning the processing of personal data and on the free movement of such data, and repealing Directive 95/46/EC (General Data Protection Regulation) \href{https://eur-lex.europa.eu/legal-content/EN/TXT/?uri=CELEX\%3A32016R0679}{https://eur-lex.europa.eu/legal-content/EN/TXT/?uri=CELEX\%3A32016R0679}} and the AI Act\footnote{Proposal for a Regulation of the European Parliament and of the Council laying down harmonized rules on artificial intelligence (Artificial Intelligence Act) and amending certain Union legislative acts \href{https://eur-lex.europa.eu/legal-content/EN/TXT/?uri=CELEX\%3A52021PC0206}{https://eur-lex.europa.eu/legal-content/EN/TXT/?uri=CELEX\%3A52021PC0206}} imposed sanctions and restrictions on the use of such models due to their unexplainable nature.

In recent years, explainable artificial intelligence (XAI) tools have been employed to open the black boxes, in other words, to explain opaque models \cite{arrieta_et_al_2020,samek_et_al_2019}. These tools have rapidly developed to solve the uninterpretable model structures in finance. The preliminary work by \cite{momparler_et_al_2016} investigated the importance of the variables in the bank failure prediction models, and then \cite{petropoulos_et_al_2020} obtained the most important variables similarly. \cite{park_et_al_2021} explains their financial distress models for the companies to obtain the important variables that can be used to treat loan eligibility requirements. \cite{zhang_et_al_2022} explains the financial distress model to meet the needs of external users generated by local and global explanations. \cite{liu_et_al_2023} explain the models that are used to predict the financial distress of the companies in China to ensure the reliability of model outputs. While there are studies on explaining black box models in finance, studies on bank failure prediction models have focused only on variable importance. Therefore, it is clear that there is a need for research on explainability in terms of understanding and accountability of such models. Not only have the explainable models, but it is also utilized to derive actionable insights from the models \cite{karimi_et_al_2020,mothilal_et_al_2020}: counterfactual explanations (CE), which describe the changes in the variable values needed to flip the prediction in the intended direction, are used to be actionable. The CE is one of the popular and easy-to-understand XAI tools in many domains such as healthcare \cite{tanyel_et_al_2023}, education \cite{tsiakmaki_et_al_2021,zhang_et_al_2023}, finance \cite{grath_et_al_2018,dastile_et_al_2022,wang_et_al_2023,cho_and_shin_2023,gunonu_et_al_2024}. It is an output of an optimization process and is generated with methods such as What-If \cite{wexler_et_al_2019}, Nearest instance \cite{brughmans_et_al_2023}, and Multi-objective counterfactual explanation \cite{dandl_et_al_2020}. The number of CEs that can be generated is potentially limitless \cite{byrne_2019}; however, generating high-quality CEs can be very difficult \cite{artelt_and_hammer_2019} because they are quite rare in most cases \cite{keane_and_symth_2020}. The high-quality CEs should be plausible, proximity, sparse, and valid \cite{keane_and_symth_2020}. There are preliminary studies to explore the desired properties of CEs in terms of desiderata. \cite{smyth_and_keane_2022} indicated that the quality counterfactuals should be sparse with two variable changes from a psychological point of view. \cite{artelt_et_al_2021} proposed using more plausible CEs instead of proximity CEs because the plausible CEs are more robust. However, no uniformly better CE generation method exists for all datasets from different domains \cite{dandl_et_al_2023}. Thus, it is necessary to evaluate quantitatively \cite{hvilshoj_et_al_2021,ge_et_al_2021} the generated CEs by different methods before deployment.

Additionally, the challenges encountered during the model selection phase include poor out-of-time performance and the issue of imbalanced data. The diminished out-of-time performance of bank failure prediction models results in their swift obsolescence \cite{du_et_al_2011,manthoulis_et_al_2020}. Hence, besides training models with robust out-of-sample performance, ensuring high out-of-time performance for long-term horizon \cite{audrino_et_al_2019,petropoulos_et_al_2020,gunonu_et_al_2023} is also imperative. Conversely, the issue of imbalanced data surfaces when there's a notable scarcity of data examples for failed banks, leading to misclassification by the bank failure prediction model, particularly for the minority class. From the data-centric AI perspective, various resampling methods such as oversampling, undersampling, and Synthetic Minority Over-sampling Technique \cite{chawla_et_al_2002} are employed to mitigate this. However, the efficacy of resampling methods has come under scrutiny in recent years due to their adverse effects on the model \cite{van_et_al_2022,junior_and_pisani_2022,stando_et_al_2024,cavus_et_al_2024,carriero_et_al_2024}, sparking debates regarding their utility \cite{elor_et_al_2022}. Thus, we lean towards using the cost-sensitive approach rather than the resampling methods to address the imbalanced data. Consequently, the impact of resampling techniques on models has been rigorously assessed on both out-of-sample and out-of-time datasets, compared with results obtained using cost-sensitive approaches.

Support vector machines, neural networks, and discriminant analysis are frequently enlisted in bank failure prediction modeling. However, despite their established efficacy in tabular datasets, tree-based models have not garnered widespread adoption, in contrast to deep learning networks \cite{carmona_et_al_2019,petropoulos_et_al_2020,grinsztajn_et_al_2022}. Hence, we offer a novel perspective by leveraging tree-based models, namely decision trees \cite{breiman_et_al_1984}, random forests \cite{breiman_2001}, and extra trees \cite{geurts_et_al_2006}. The rationale behind employing multiple models lies in their diverse predictive capabilities, yielding different outcomes with varying variances. This paper solves basic bank failure prediction problems such as imbalance, out-of-sample, and out-of-time performance for US banks. Then, it determines the most appropriate method to generate actionable explanations from black-box bank failure prediction models, considering the effects of the balancing strategies used to produce the highest quality counterfactual explanations. To our knowledge, this is the first paper that proposes using counterfactual explanations to extract actionable insights from the bank failure prediction models. This provides valuable insights to reduce the failure risk of banks, in addition to understanding the model. Moreover, we solve the problems of imbalancedness and out-of-time performance while obtaining the desired counterfactuals. In summary, our contribution is: (1) to train an accurate bank failure prediction model for the US banks against addressing imbalanced data and long time horizon, (2) to obtain the best counterfactual generation method for the bank failure prediction model, and (3) to illustrate how to use the counterfactual explanations to avoid bank failure in practice. This paper is organized as follows. Section~\ref{sec:methods} provides an overview of the counterfactual explanations, and Section~\ref{sec:experiments} includes information about the conducted experiments on counterfactual generation methods and presents the results. The last section summarises the findings 
in the paper and suggests further studies.

\section{Methods}
\label{sec:methods}
This section defines counterfactual explanations, introduces their desiderata, and provides the generation methods and evaluations of counterfactual explanations.

\subsection{Counterfactual Explanations}
A counterfactual explanation (CE) is the changes that must be made to certain variables to achieve the desired level of model prediction. In other words, a CE shows the action that can be taken to alter the model output. Let $D = \{(x_i, y_i)\}_{i=1}^n$ represent a dataset, where $x_i$ denotes a $d$-dimensional variable vector in $\mathbb{R}^d$, and $y_i \in [0, 1]$ represents the target variable. Each sample $(x_i, y_i)$ is drawn independently from a joint distribution. The objective of a binary classifier is to learn an optimal mapping function  $f_D: X \rightarrow Y$ by minimizing a loss function defined as $L(f) = \mathbb{P}[Y \neq f_D(X)]$. A factual observation $x \in X$, and a counterfactual observation $x^\prime \in X$ that has the closest observation to $x$ and has the opposite target. This process can be seen as an optimization problem achieved to the desired outcome. However, the number of solutions, i.e., counterfactual explanations, is potentially limitless \cite{byrne_2019} and they have different properties in terms of their desiderata such as validity, proximity, sparsity, and plausibility (please refer to the paper of \cite{guidotti_2022} for more detailed information about the CEs). These properties are introduced in the following section.

\subsection{Desiderata for counterfactual explanations}
The CEs to achieve the desired outcome can be described in terms of several properties \cite{verma_et_al_2020}. However, we consider the basic properties \textit{validity, proximity, sparsity, and plausibility} that are introduced as follows.\\

\noindent \textbf{\textbf{Validity}} refers to the extent to which a CE accurately reflects the true state of affairs, demonstrating the correct causal factors underlying the given outcome and accurately identifying the factors influencing the likelihood of occurrence. Let $x$ represent an observation and $x^\prime$ denote the counterfactual observation. The validity of a CE $x^\prime$ can be formulated as $\text{validity}(x^\prime) = \mathbb{I}\{ d(x, x^\prime) \leq \epsilon$ where $d(x, x^\prime)$ represents the distance between the observation $x$ and the counterfactual observation $x^\prime$, and $\epsilon$ is a predefined threshold that determines the acceptable level of dissimilarity. $\mathbb{I}\{\cdot\}$ denotes the indicator function, returning $1$ if the condition inside the brackets is true and $0$ otherwise.
\\

\noindent \textbf{\textbf{Proximity}} indicates how closely a CE aligns with the given outcome. An achievable CE should closely approximate the actual situation, allowing for the evaluation of the similarities between alternative scenarios and the actual state of affairs. Proximity measures how close a counterfactual observation $x^\prime$ is to the observation $x$. It can be defined as $\text{proximity}(x, x^\prime) = d(x, x^\prime)$ where $d(x, x^\prime)$ represents the distance metric between the observation $x$ and the counterfactual observation $x^\prime$. 
\\

\noindent \textbf{\textbf{Sparsity}} assesses the complexity of a CE. A high-quality explanation should be concise and free from unnecessary information, providing a clear and straightforward representation. Complexity or excessive detail in explanations may hinder comprehension. Sparsity evaluates the complexity of a CE $x^\prime$ in terms of the number of variables involved. It can be quantified as $\text{sparsity}(x^\prime) = \text{c}(x^\prime)$ where $\text{c}(x^\prime)$ represents the number of variables needed to change in the counterfactual observations $x^\prime$.
\\

\noindent \textbf{\textbf{Plausibility}} refers to the extent to which a CE is logical and realistic. It should be consistent with general knowledge and facts, aligning with established principles and reflecting real-world feasibility. Explanations lacking plausibility may undermine user confidence and hinder the formation of an accurate understanding. Plausibility assesses the logical coherence and realism of a CE $x^\prime$. It can be expressed as $\text{plausibility}(x^\prime) = \mathbb{P}(x^\prime)$ where $\mathbb{P}(x^\prime)$ represents the probability or likelihood of observing the counterfactual observation $x^\prime$ given the observed data and context.

\subsection{Counterfactual generation methods}
Over one hundred and twenty methods for counterfactual generation are proposed \cite{warren_et_al_2023}. These methods differ in terms of the structure of the objective functions they target and, therefore, the characteristics of the CEs. In this section, we briefly introduce three generation methods that we focused on. Considering all methods in the literature is not feasible because of the number of available methods. 

\subsubsection{Multi-objective counterfactual explanations} Multi-objective counterfactual explanation (MOC) method handles the counterfactual search work as a multi-objective optimization problem \cite{dandl_et_al_2020}. It focuses on generating \textbf{proximit}, \textbf{valid}, \textbf{plausible}, and \textbf{sparse} counterfactual explanations, thus trying to solve the following optimization problem:
\begin{equation}
    min_x o(x) = min_x [o_1(\hat{f}(x), Y^\prime), o_2(x, x^\prime), o_3(x, x^\prime), o_4(x, X)]
\end{equation}
\noindent MOC employs a customized version of the NSGA-II algorithm, integrating mixed integer evolutionary strategies, a distinct crowding distance sorting method, and optional adaptations designed for the counterfactual search process.

\subsubsection{What If counterfactual explanations} The What-if method (WhatIf) identifies observations in proximity to a given observation $x$ from a set of other observations, based on the Gower distance metric, by solving the optimization problem formulated as follows \cite{wexler_et_al_2019}: 
\begin{equation}
    x^\prime \in \text{argmin}_{x \in X} d(x, x^\prime).
\end{equation}

\subsubsection{Nearest instance counterfactual explanations}
The Nearest instance counterfactual explanations (NICE) method adopts the Heterogeneous Euclidean Overlap Method to measure the distance between the observation $x$ and the counterfactual $x^\prime$ \cite{brughmans_et_al_2023}. For each variable $F$, the distance $d_F(a, b)$ between two variable values $a$ and $b$ is calculated as follows:
\begin{equation}
d_F(a, b) =
\begin{cases}
1 & \text{if } a \neq b \text{ for categorical } F \\
0 & \text{if } a = b \text{ for categorical } F \\
\frac{|a - b|}{\text{range}(F)} & \text{for numerical } F
\end{cases}
\end{equation}

\noindent where $\text{range}(F)$ represents the range of values of variable $F$. The total distance $d(x_0, x_c)$ is computed as the L1-norm of all variable distances. Furthermore, this metric ensures that the contribution of each variable to the total distance is between 0 and 1.

\section{Experiments}
\label{sec:experiments}
In this section, we first train an accurate classification model for predicting bank failures. Then, we find the best counterfactual generation method for the most accurate bank failure prediction model in terms of the desiderata of the counterfactual explanations. Hence, the dataset and the variables are introduced in the models, the modeling phase is detailed, and lastly, the quality performance of the counterfactuals is compared.\\

\noindent \textbf{Dataset.} We collected the data about failed and non-failed banks in the US using the \texttt{R} package \texttt{fdicdata}, which is connected to the Federal Deposit Insurance Corporation database \cite{fdicdata}. While the number of observations in active banks is $445.493$, it is $39.835$ in closed banks. We used the dataset covering $15$ years between $2008$ and $2023$. It is split into two parts because the bank failure prediction model works accurately in out-of-sample and out-of-time datasets. Thus, the data between $2008-2014$ is used as in-sample and out-of-sample, and $2014-2023$ is used as out-of-time. Also, the holdout validation is used in the out-of-sample performance of the model, $80\%$ of the data set as an in-sample and $20\%$ as an out-of-sample of the data between $2008-2014$. The models were trained on three predictor groups, I, II, and III, including the combinations of CAMELS (Capital, Asset Quality, Management Adequacy, Earnings, Liquidity, and Sensitivity to Market Risk) indicators. These predictor groups are taken from the literature papers. The predictors I and II are used in \cite{gogas_et_al_2018} and \cite{petropoulos_et_al_2020}, respectively. Predictors III, which consists of different indicators, is also proposed in this paper. The predictors in the predictor groups are listed in detail in Table~\ref{tab:variables}.

\begin{table}
    \centering
    \caption{Details of variables used in the data set}
    \label{tab:variables}
    \begin{tabular}{p{1.6cm}p{2cm}p{2.3cm}p{6cm}} \toprule
             Predictors & Indicator & Range             & Description\\\toprule
     \multirow{4}{*}{I} &  \texttt{TICRC}    & $[-0.01, 0.19]$   & Tier 1 Risk-Based Capital Ratio / Total Assets\\
                        &  \texttt{PLLL}     & $[-3, 10]$        & Provisions for Loan \& Lease Losses / Total Interest Income\\
                        &  \texttt{TIE}      & $[0, 2.2]$        & Total Interest Expense / Total Interest Income\\
                        &  \texttt{EQR}      & $[-20, 100]$      & Equity Capital Ratio\\\midrule
    \multirow{4}{*}{II} &  \texttt{TICRC}    & $[-0.01, 0.19]$   & Tier 1 Risk-Based Capital Ratio / Total Assets\\
                        &  \texttt{NIMY}     & $[-4, 26]$        & Net Interest Margin\\
                        &  \texttt{INTEXPYQ} & $[-0.5, 5.5]$     & Cost of Funding Earning Assets Quarterly\\
                        &  \texttt{RBCIAAJ}  & $[-20, 200]$      & Leverage Ratio\\
                        &  \texttt{ROE}      & $[-12000, 1000]$  & Return on Equity\\\midrule
    \multirow{4}{*}{III}&  \texttt{NIMYQ}    & $[-4, 26]$        & Net Interest Margin Quarterly\\
                        &  \texttt{LNATRESR} & $[0, 26]$         & Loan Loss Reserve / Gross Loan \& Lease\\
                        &  \texttt{NONIXAYQ} & $[-20, 300]$      & Noninterest Expenses / Average Assets Quarterly\\
                        &  \texttt{ROAQ}     & $[-100, 350]$     & Quarterly Return on Assets\\\bottomrule
    \end{tabular}
\end{table}
\noindent \textbf{Modeling.} We employed tree-based predictive models, including decision trees, random forests, and extra trees, each applied to three distinct sets of variables as delineated in Table~\ref{tab:perf1}. Notably, all predictions were derived without employing any resampling techniques. However, subsequent analyses incorporated different resampling methods such as undersampling, oversampling, Synthetic Minority Over-sampling Technique (SMOTE), and the cost-sensitive approach tailored to each model. We split the dataset into in-sample, out-of-sample, and out-of-time segments for evaluation purposes. Additionally, the data preprocessing stage specifically considered a 1-year lag period for closed banks. Prediction outcomes were then extrapolated for the out-of-time segment, covering the period from $2014$ to $2023$. Across all models utilized, default parameter settings were retained, with resampling methods applied after assessing and addressing any imbalances observed in the training and validation datasets. We effectively mitigated any imbalances in the classification task by adopting resampling methods.\\

\noindent \textbf{Model performance comparisons.} Performance of the trained models is evaluated in terms of accuracy and F1 score. We used the F1 score because of the imbalanced distribution of the failed and non-failed banks. It is calculated over the harmonic mean of precision and recall \cite{lipton_et_al_2014} and provides additional insights into model performance. Detailed results of our analyses, encompassing all predictor groups and models employed, are presented comprehensively in Tables~\ref{tab:perf1} and \ref{tab:perf2}.

The results presented in Table~\ref{tab:perf1} show that the model trained on the original dataset and the model trained with the cost-sensitive approach exhibit superior accuracy and F1 scores. Predictor I, with the Decision Tree model trained on the original dataset, has an accuracy of 0.9784 and an F1 score of 0.8981. In contrast, the model trained with the cost-sensitive approach has an accuracy of 0.9381 and an F1 score of 0.7860, underscoring the efficacy of the cost-sensitive approach in addressing data imbalance. Conversely, the model trained on undersampled data generally yields the lowest performance; for the same model and predictor, the accuracy is 0.9558, and the F1 score is 0.8207, suggesting that this way leads to information loss, consequently diminishing performance.

Comparatively, the Extra Trees and Random Forest models consistently outperform the Decision Tree model regarding higher accuracy and F1 scores. For example, Predictor I with the Extra Trees model trained on the original dataset achieves an accuracy of 0.9872 and an F1 score of 0.9372, and the model trained with the cost-sensitive approach achieves an accuracy of 0.9882 and an F1 score of 0.9423. Similarly, the Random Forest model demonstrates robust performance, trained on the original dataset, achieving an accuracy of 0.9862 and an F1 score of 0.9333, and the model trained with the cost-sensitive approach achieves an accuracy of 0.9872 and an F1 score of 0.9383. This indicates that more complex and ensemble-based models are more effective in predicting bank failures.

Regarding predictors, the models with Predictor II generally achieve the highest accuracy and F1 scores, suggesting its preference for achieving higher performance in bank failure prediction models. For instance, with the Decision Tree model trained on the original dataset, Predictor II achieves an accuracy of 0.9852 and an F1 score of 0.9282. In contrast, the model trained with the cost-sensitive approach has an accuracy of 0.9803 and an F1 score of 0.9090. Similar high performance is observed for the Extra Trees and Random Forest models. Overall, the models trained on the original datasets and the model trained with the cost-sensitive approach provide the best performance. In contrast, the undersampling method reduces performance, and the Extra Trees and Random Forest models exhibit superior performance. The utilization of Predictor II can further enhance model performance on out-of-sample.

\begin{sidewaystable}
    \centering
    \caption{Accuracy and F1 values for out-of-sample with three different models for predictors}
    \begin{tabular}{cccccccccccc}\toprule
         & & \multicolumn{5}{c}{Accuracy} & \multicolumn{5}{c}{F1}\\\midrule
        Model & Predictors & \rotatebox{90}{Original} & \rotatebox{90}{Undersampling} & \rotatebox{90}{Oversampling} & \rotatebox{90}{SMOTE} & \rotatebox{90}{Cost-sensitive} & \rotatebox{90}{Original} & \rotatebox{90}{Undersampling} & \rotatebox{90}{Oversampling} & \rotatebox{90}{SMOTE} & \rotatebox{90}{Cost-sensitive} \\\midrule
                        & I     & \textbf{0.9784} & 0.9558 & 0.9577 & 0.9735 & 0.9705 & \textbf{0.8981} & 0.8207 & 0.8273 & 0.8831 & 0.8672\\
         Decision tree  & II    & \textbf{0.9852} & 0.9823 & 0.9665 & 0.9626 & 0.9813 & \textbf{0.9282} & 0.9158 & 0.8521 & 0.8430 & 0.9132\\
                        & III   & \textbf{0.9577} & 0.8928 & 0.8968 & 0.8938 & 0.9381 & \textbf{0.7860} & 0.6472 & 0.6579 & 0.6516 & 0.7449\\\midrule
                        & I     & \textbf{0.9872} & 0.9764 & 0.9862 & 0.9813 & 0.9843 & \textbf{0.9372} & 0.8928 & 0.9326 & 0.9116 & 0.9245\\
         Extra trees    & II    & 0.9872 & 0.9842 & 0.9882 & 0.9823 & \textbf{0.9882} & 0.9372 & 0.9259 & 0.9423 & 0.9174 & \textbf{0.9423}\\
                        & III   & \textbf{0.9607} & 0.9390 & 0.9607 & 0.9518 & 0.9587 & 0.8148 & 0.7596 & \textbf{0.8198} & 0.7967 & 0.8157\\\midrule
                        & I     & 0.9862 & 0.9784 & 0.9853 & 0.9803 & \textbf{0.9862} & 0.9333 & 0.9027 & 0.9282 & 0.9082 & \textbf{0.9333}\\
         Random forest  & II    & 0.9872 & 0.9803 & 0.9872 & 0.9823 & \textbf{0.9872} & 0.9378 & 0.9090 & 0.9378 & 0.9174 & \textbf{0.9383}\\
                        & III   & \textbf{0.9626} & 0.9479 & 0.9626 & 0.9518 & 0.9607 & 0.8224 & 0.7871 & \textbf{0.8303} & 0.7967 &0.8245\\\bottomrule
    \end{tabular}
    \label{tab:perf1}
\end{sidewaystable}


The results presented in Table~\ref{tab:perf2} show that the models trained on the original dataset and those trained with the cost-sensitive approach exhibit superior accuracy and F1 scores. This is particularly significant when interpreting the performance of various models across different sampling methods. For instance, Predictor I with the Decision Tree model, which is trained on the original dataset, has an accuracy of 0.9936 and an F1 score of 0.4000. In contrast, the model trained with the cost-sensitive approach attains an accuracy of 0.9957 and an F1 score of 0.7500, indicating the effectiveness of the cost-sensitive approach in addressing data imbalance. Conversely, the oversampling method generally yields the lowest performance; for the same model and predictor, the oversampling accuracy is 0.9700, and the F1 score is 0.3333, suggesting this method may be less effective.

Comparatively, the Extra Trees and Random Forest models consistently outperform the Decision Tree model regarding higher accuracy and F1 scores. For example, Predictor II with the Extra Trees model trained on the original dataset achieves an accuracy of 0.9989 and an F1 score of 0.9230, and the model trained with the cost-sensitive approach also achieves an accuracy of 0.9989 and an F1 score of 0.9230. Similarly, the Random Forest model demonstrates robust performance, which is trained on the original dataset, achieving an accuracy of 0.9957 and an F1 score of 0.6666, and trained with the cost-sensitive approach, achieving an accuracy of 0.9968 and an F1 score of 0.8000. This indicates that the models trained with the cost-sensitive approach are more effective in predicting bank failures.

Regarding predictors, the models trained with Predictor II generally achieve the highest accuracy and F1 scores, suggesting its preference for achieving higher performance in bank failure prediction models. For instance, Predictor II, with the Decision Tree model trained on the original datasets and also trained with the cost-sensitive approach, achieved an accuracy of 0.9989 and an F1 score of 0.9230. Similar high performance is observed for the Extra Trees and Random Forest models. Overall, the models trained on the original dataset and those trained with the cost-sensitive approach provide the best results. In contrast, the oversampling method reduces performance, and the Extra Trees and Random Forest models exhibit superior performance. Using Predictor II can further enhance the model's accuracy for out-of-time performance.


\begin{sidewaystable}
    \centering
    \caption{Accuracy and F1 values for out-of-time with three different models for predictors}
    \begin{tabular}{cccccccccccc}\toprule
         & & \multicolumn{5}{c}{Accuracy} & \multicolumn{5}{c}{F1}\\\midrule
        Model & Predictors & \rotatebox{90}{Original} & \rotatebox{90}{Undersampling} & \rotatebox{90}{Oversampling} & \rotatebox{90}{SMOTE} & \rotatebox{90}{Cost-sensitive} & \rotatebox{90}{Original} & \rotatebox{90}{Undersampling} & \rotatebox{90}{Oversampling} & \rotatebox{90}{SMOTE} & \rotatebox{90}{Cost-sensitive} \\\midrule
                        & I     & 0.9936 & 0.9689 & 0.9700 & 0.9592 & \textbf{0.9957} & 0.4000 & 0.3255 & 0.3333 & 0.2692 & \textbf{0.7500}\\
         Decision tree  & II    & 0.9989 & 0.9989 & 0.9602 & 0.9677 & \textbf{0.9989} & 0.9230 & 0.9230 & 0.2448 & 0.2857 & \textbf{0.9230}\\
                        & III   & 0.9946 & 0.9571 & 0.9356 & 0.9635 & \textbf{0.9968} & 0.5454 & 0.2307 & 0.1666 & 0.2608 & \textbf{0.7692}\\\midrule
                        & I     & 0.9946 & 0.9356 & 0.9345 & 0.9871 & \textbf{0.9957} & 0.5454 & 0.1891 & 0.1866 & 0.5384 & \textbf{0.6666}\\
         Extra trees    & II    & 0.9989 & 0.9828 & 0.9989 & 0.9925 & \textbf{0.9989} & 0.9230 & 0.4285 & 0.9230 & 0.6315 & \textbf{0.9230}\\
                        & III   & 0.9946 & 0.9614 & 0.9946 & 0.9861 & \textbf{0.9946} & 0.5454 & 0.2800 & 0.5454 & 0.4347 & \textbf{0.5454}\\\midrule
                        & I     & 0.9957 & 0.9356 & \textbf{0.9979} & 0.9871 & 0.9968 & 0.6666 & 0.1891 & 0.8571 & 0.5384 & \textbf{0.8000}\\
         Random forest  & II    & 0.9989 & 0.9696 & 0.9989 & 0.9925 & \textbf{0.9989} & 0.9230 & 0.3000 & 0.9230 & 0.6315 & \textbf{0.9230}\\
                        & III   & 0.9946 & 0.9485 & 0.9957 & 0.9861 & \textbf{0.9957} & 0.5454 & 0.2000 & 0.6666 & 0.4347 & \textbf{0.6666}\\\bottomrule
    \end{tabular}
    \label{tab:perf2}
\end{sidewaystable}

Examining the results from Table~\ref{tab:perf1}, which represents models' out-of-sample performance, and Table~\ref{tab:perf2}, which illustrates their out-of-time performance, provides valuable insights into the effectiveness of different models, predictors, and sampling methods in predicting bank failures. In Table~\ref{tab:perf1}, focusing on out-of-sample performance, Predictor II consistently emerges as the top performer across various models and datasets. For example, using the Extra Trees model, Predictor II achieves high accuracies of up to 0.9872 and F1 scores of 0.9372 with the cost-sensitive approach. This indicates Predictor II's robust capability to generalize well to new, unseen data, which is crucial for reliable model deployment.

In contrast, Table~\ref{tab:perf2}, showcasing out-of-time performance, emphasizes the importance of models that can maintain accuracy over time despite potential shifts in data characteristics. Here again, Predictor II demonstrates superior performance, achieving accuracies of 0.9989 and F1 scores of 0.9230 using the Extra Trees model in the original and cost-sensitive datasets. This consistency underscores Predictor II's reliability in predicting bank failures even when confronted with new data from different time horizons, highlighting its resilience to temporal variations.

Overall, while the results in Tables~\ref{tab:perf1} and \ref{tab:perf2} underscore the effectiveness of Extra Trees and Random Forests, Predictor II stands out for its ability to deliver high performance across out-of-sample and out-of-time scenarios. This suggests that Predictor II, coupled with Random Forest, enhances predictive accuracy and ensures robustness in forecasting bank failures under varying conditions. Thus, for bank failure prediction in the US, Predictor II with Random Forest models emerges as the preferred choice due to its superior performance in both out-of-sample and out-of-time contexts.\\

\begin{figure}[H]
    \centering
    \includegraphics[angle=90, scale = 0.27]{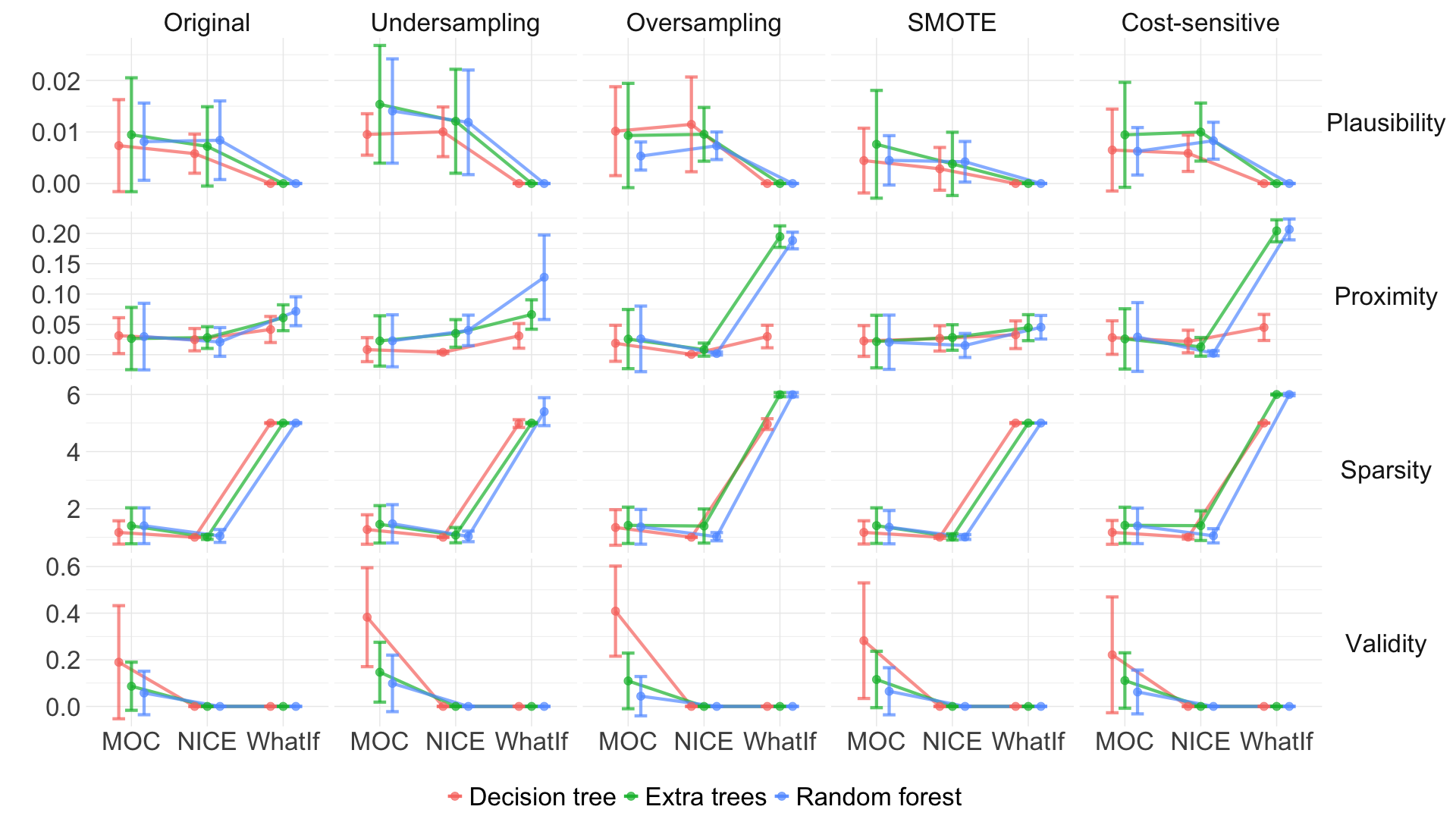}
    \caption{Mean and standard deviations plot of the counterfactual properties for the methods and resampling strategies}
    \label{fig:CE}
\end{figure}

\noindent \textbf{CE comparison.} We compared counterfactual generation methods MOC, NICE, and WhatIf in terms of desiderata \textit{plausibility, proximity, sparsity}, and \textit{validity}, showing how their performance varies depending on resampling methods and the cost-sensitive approach to handle the imbalancedness and tree-based machine learning models Decision tree, Extra trees, and Random forest. Four resampling methods, Undersampling, Oversampling, and SMOTE, are used with the cost-sensitive approach and the Original data, which is not pre-processed for balancing as a baseline. The results are given in Figure~\ref{fig:CE} in terms of the mean values of the desiderata with the standard deviation values in the error bars. The lower values are better in each metric, and the length of the error bar shows the values' instability.

We evaluated the effect of resampling methods and the cost-sensitive approach on two dimensions: CE generation methods and models. Firstly, the results are compared over resampling methods. In terms of plausibility, generally, low values are observed across all resampling methods, indicating that the CEs are of high quality in terms of plausibility. The Extra Trees model typically stands out with lower plausibility values. While the CEs generated on the models trained on the original dataset show a broader range of plausibility values, the CEs generated on the resampled dataset show narrower ranges. The NICE method usually has lower proximity values, indicating that CEs are closer to the real data. However, in the SMOTE and cost-sensitive method, the proximity values show higher variability, suggesting a broader instability in some cases. The MOC and NICE methods generate sparser CEs, and the WhatIf generates worse CEs in terms of sparsity. Particularly with the undersampling and oversampling methods, there are sparser explanations. When examining the variability of these metrics, again, the CEs generated on the models trained on the original dataset have more consistent sparsity values, whereas other methods display more significant variability. This finding can be interpreted as the resampling methods generating the CEs with unstable sparsity. The WhatIf method shows lower values for the validity metric, especially in the SMOTE and cost-sensitive methods, providing more valid explanations. However, validity values vary widely in these two methods, indicating high variability.

The MOC and NICE methods demonstrate superior plausibility, proximity, and sparsity metrics, offering desired CEs. The cost-sensitive technique performs better across many metrics, making it advantageous for generating more desired CEs. Regarding variability, the CEs generated on the models trained on the original dataset have less variability in metric values, whereas other resampling methods show a broader range of performance values. This means that the resampling methods lead to unstable CEs being generated. As a result, we propose using NICE methods to generate higher-quality CEs with the cost-sensitive method used to handle the imbalance.\\


\section{Applications}
In this section, we illustrate how counterfactual explanations can be used to take action by banks to flip the case of failure. We used the \textbf{Random forest} model trained with the \textbf{cost-sensitive} approach using \textbf{Predictors II} and utilized the NICE method to generate the counterfactuals based on the findings in the previous section. In the applications, we randomly select two banks that are likely to fail according to our bank failure prediction model. We name these two banks, A and B, to ensure data privacy. The observation and the counterfactual of Banks A and B are given in Tables~\ref{tab:CE_bankA} and \ref{tab:CE_bankB}, respectively. 

Table~\ref{tab:CE_bankA} shows the NICE method generates one counterfactual for Bank A. It means there is only one way to change the situation of failure. The $x^\prime_A$ indicates that the values of the variable \texttt{NIMY} (Net Interest Margin) should be decreased. In contrast, the values of the variables \texttt{TICRC} (Tier 1 Risk-Based Capital Ratio / Total Assets), \texttt{RBCIAAJ} (Leverage Ratio) and \texttt{ROE} (Return on Equity) should be increased. The risk of failure can be flipped if the way pointed out by $x^\prime_A$ followed. 

\begin{table}
    \centering
    \caption{The counterfactuals generated by NICE method for Bank A}
    \begin{tabular}{crrrrr}\hline
                      & \texttt{TICRC}              & \texttt{NIMY}         & \texttt{INTEXPYQ} & \texttt{RBCIAAJ}       & \texttt{ROE}           \\\hline
        $x_A$         & 2.884209e-06                & 3.2700                & 2.65              & 7.62981                & -20.47                 \\
        $x^\prime_A$  & $\uparrow$ 2.097704e-04     & $\downarrow$ 0.8900   & 2.65              & $\uparrow$ 8.04720     & $\uparrow $1.72        \\\hline
    \end{tabular}
    \label{tab:CE_bankA}
\end{table}

Table~\ref{tab:CE_bankB} shows the NICE method generates three counterfactuals for Bank B. It means there are several ways to change the situation of failure. The $x^\prime_{B1}$ indicates that the value of the variable \texttt{INTEXPYQ} (Cost of Funding Earning Assets Quarterly) should be decreased while the value of the variable \texttt{RBCIAAJ} should be increased. The other counterfactuals show the different ways to reduce the risk, such as the $x^\prime_{B2}$ indicates that the value of the variable \texttt{NIMY} should be decreased. In contrast, the value of the variable \texttt{RBCIAAJ} should be increased, and the $x^\prime_{B3}$ indicates that the value of the variable \texttt{TICRC} should be decreased. In contrast, the value of the variable \texttt{RBCIAAJ} should be increased. All counterfactuals suggest increasing the value of the variable \texttt{RBCIAAJ} while decreasing the values of different variables. The risk of failure can be flipped if one of the ways pointed out by $x^\prime_A$ followed. 

\begin{table}
    \centering
    \caption{The counterfactuals generated by NICE method for Bank B}
    \begin{tabular}{crrrrr}\hline
                            & \texttt{TICRC}            & \texttt{NIMY}         & \texttt{INTEXPYQ}     & \texttt{RBCIAAJ}      & \texttt{ROE}  \\\hline
        $x_B$               & 3.838692e-06              & 3.148666              & 2.47                  & 11.37229              & 4.63          \\
        $x^\prime_{B1}$     & 3.838692e-06              & 3.148666              & $\downarrow$ 2.07     & $\uparrow$ 11.66318   & 4.63          \\
        $x^\prime_{B2}$     & 3.838692e-06              & $\downarrow$ 2.767765 & 2.47                  & $\uparrow$ 11.66318   & 4.63          \\
        $x^\prime_{B3}$     & $\downarrow$ 2.314775e-05 & 3.148666              & 2.47                  & $\uparrow$ 11.66318   & 4.63          \\\hline
    \end{tabular}
    \label{tab:CE_bankB}
\end{table}

Unlike Bank A, the NICE method produces more than one counterfactual explanation for Bank B. This provides selection alternatives for the user, who can use the desired description. On the other hand, this may also impose a burden of choice on the user. In this case, the user can choose by considering the field dynamics or calculating the desiderata for each explanation and choose with the help of these metrics. As seen in applications, CEs can produce recommendations for banks likely to fail, using the trained model to reduce their risk of failure.

\section{Conclusions}

This paper investigates the efficacy of various counterfactual explanation methods in predicting bank failures using tree-based machine-learning models. The research highlights the critical importance of generating high-quality counterfactual explanations, focusing on key metrics such as plausibility, proximity, sparsity, and validity. The findings indicate that the Multi-Objective Counterfactual (MOC) and Nearest Instance Counterfactual Explanations (NICE) methods outperform other techniques in delivering high-quality counterfactual explanations, with NICE particularly excelling in proximity and sparsity metrics when used with the cost-sensitive approach.

The empirical analysis demonstrates that different resampling techniques, such as undersampling, oversampling, and SMOTE, significantly impact the performance of counterfactual explanation methods. Specifically, the cost-sensitive approach effectively handles data imbalance, leading to more accurate and reliable counterfactual explanations. The NICE method consistently shows lower proximity values, indicating that the counterfactual explanations are closer to the real data. In comparison, WhatIf and NICE methods achieve lower sparsity values, suggesting more straightforward and more interpretable explanations.

The study also highlights the variability in the performance of counterfactual explanation methods across different tree-based models, including Decision Trees, Extra Trees, and Random Forests. Ensemble models like Random Forests and Extra Trees typically exhibit superior predictive accuracy and robustness performance. In particular, the Extra Trees model stands out for its lower plausibility values across various resampling methods, indicating high-quality counterfactual explanations in terms of plausibility.

Furthermore, the findings emphasize the stability of counterfactual explanations generated on the original dataset compared to those produced with resampling methods. The counterfactual explanations on the original dataset show more consistent values across all metrics. In contrast, those generated with resampling techniques exhibit more significant variability, highlighting the challenges in achieving stability with different resampling methods.

These results underscore the potential of leveraging advanced counterfactual explanation methods and tree-based models to enhance bank failure prediction models' predictive accuracy and interpretability. The research demonstrates that integrating counterfactual explanations with the cost-sensitive approach can effectively address data imbalance issues, providing more desirable and reliable explanations.

In the context of regulatory requirements such as the GDPR and the AI Act, the ability to generate interpretable and reliable explanations for predictive models is increasingly critical. This study contributes to the growing body of literature by providing empirical evidence on the effectiveness of counterfactual explanations in improving model transparency and trustworthiness.
\section*{Acknowledgments}
The work on this paper is financially supported by the Scientific and Technological Research Council of Turkiye under 2209-A- Research Project Support Programme for Undergraduate Students grant no. 1649B022303919 and Eskisehir Technical University Scientific Research Projects Commission under grant no. 24LÖP006.

\section*{Supplemental Materials}

The materials for reproducing experiments performed and the dataset are in the repository: \href{https://github.com/mcavs/Explainable_bank_failure_prediction_paper}{https://github.com/mcavs/Explainable\_bank\_failure\_prediction\_paper}

\end{document}